\pdfoutput=1
\documentclass[11pt]{article}
\usepackage[final]{acl}
\usepackage{newtxtext}
\usepackage{newtxmath}
\usepackage{latexsym}
\usepackage[T1]{fontenc}
\usepackage[utf8]{inputenc}
\usepackage{microtype}
\usepackage{inconsolata}
\usepackage{graphicx}
\usepackage{booktabs}
\usepackage{makecell}
\usepackage{siunitx}
\usepackage{comment}

\title{Synthetic Child-Adult Interactions Are Not Developmentally Plausible}

\author{
 \textbf{Bastian Bunzeck\textsuperscript{1}},
  \textbf{Francesca Padovani \textsuperscript{2}},
 \\
  \textsuperscript{1}Computational Linguistics, Department of Linguistics, Bielefeld University,
   \\
  \textsuperscript{2}Center for Language and Cognition (CLCG), University of Groningen,
\\
  \small{
    \textbf{Correspondence:} \href{mailto:email@domain}{bastian.bunzeck@uni-bielefeld.de}
 }
}
\usepackage{xcolor}

\begin{document}

\maketitle

\begin{abstract}
Synthetic data has become a common resource for training language models, and this trend has spilled over into research on developmentally plausible language modeling. However, it remains unclear whether LM-generated child–adult interactions actually resemble natural ones. We compare the synthetic TinyDialogues dataset to natural dialogue across a broad range of linguistic measures. While synthetic data matches natural data on the lexical level, it drastically underestimates the messy, fragmented nature of spoken interaction. To mitigate this, we propose instruction-tuning with data created from real dialogues, and show in a pilot study that a small model approximates the statistical tendencies of natural data after moderate parameter-efficient fine-tuning.
\footnote{While our current results are already meaningful in their own right, this project is still work-in-progress. We are submitting this as non-archival as we are eager for feedback from the community, especially on our novel fine-tuning approach, and general desiderata for synthetic data in BabyLM.}
\end{abstract}

\section{Introduction}

Common wisdom in machine learning states that model performance generally scales with size, both in architecture (e.g. number of parameters) and data (i.e. the amount of input to the learning algorithm). While the former is generally constrained by hardware, the latter is constrained by availability. As of 2026, virtually all frontier LMs are trained on synthetic data, data generated by LMs themselves. Interestingly, this trend has begun to spill over into communities even more constrained by data availability, namely those interested in developmentally plausible language modeling (like BabyLM, cf. \citealp{feng2024childdirected,haga2024babylm,rasanen2025pipeline,zeng2026bringing,bunzeck2026child-directed}). As real transcriptions of child-adult interactions are costly to produce, transcribe and annotate, they are often not available in sufficient quantity or quality. Further, the interaction track of BabyLM even allows synthetic input from teacher models.

LMs offer a quick and easy alternative to the creation of virtually infinite amounts of similar data. However, it remains questionable if the linguistic make-up of such data resembles patterns found in natural language, which is a necessary criterion for the validity of experiments based on them. Contemporary studies show that this is not necessarily the case, e.g. on the syntactic level \cite{ju2025domain} or on the lexical and discourse levels \cite{liu2024benchmarking}, and heavily relies on context \cite{nieth2026how}. Unfortunately, no current studies analyze synthetic data broadly across different levels of linguistic description. We address this gap by comparing the synthetic dialogue from the TinyDialogues dataset \cite{feng2024childdirected} to different types of dialogue between children and caregivers, in classroom settings and between adult friends. We find that synthetic data matches natural data on the lexical level, but drastically underestimates the messy, fragmented nature of spoken data on the syntactic and construction level. It features less grammatical errors and more semantic alignment. 

To mitigate this problem, we finally propose a novel fine-tuning approach with instruction-tuning data generated from real dialogues, and show in a pilot study that even a small, 4B-parameter model can approximate the statistical tendencies of natural data after a moderate amount of parameter-efficient fine-tuning.



\section{Related work}



\paragraph{Stochastically generated child-caregiver interactions}

\citet{feng2024childdirected} introduced the TinyDialogues (TD) dataset, created by prompting GPT-4 to generate child-caregiver dialogues that contain specific words and are appropriate for specific age groups (2, 5, 10, 15). GPT-2 models trained on this type of data compare similarly on linguistic benchmarks as CHILDES-trained models, whereas TD-trained BERT models considerably outperform CHILDES models. They do not discuss possible linguistic differences between these datasets, apart from noting that the individual speaker distributions are severely mismatched (45.7\% mother vs. 38.2\% child for CHILDES, 46.6\% child vs. 8.4\% mother for TD). \citet{zeng2026bringing} extend TD by prompting GPT-4 to translate it into Spanish and generating code-switched sentences, but without testing for any kind of linguistic plausibility.

In contrast to the naive prompting approach used for TD, 
\citet{rasanen2024agedependent,rasanen2025pipeline} propose the GILES pipeline for creating more developmentally plausible pretraining data. They train a GPT-2 architecture on adult speech found in CHILDES. 
Crucially, their model is used to generate only child-directed speech (CDS), and not children's responses or any type of interaction. For this limited domain, however, the authors succeed in generating data that aligns with age-dependent patterns in natural data, e.g. for utterance length or the proportions of different parts-of-speech. Only for type-token ratio and perplexity under a reference LM they find that natural data is generally more complex. 

We acknowledge that there is also a growing body of work on generating simple (sometimes child-appropriate) stories and their linguistic properties, which lies out of the scope of our current investigation (cf. \citealp{valentini2023automatic,eldan2023tinystories,finke2025parameterized,lee2025readability}).




\paragraph{Evaluating the linguistic properties of synthetic data}

Apart from the findings by \citet{rasanen2024agedependent,rasanen2025pipeline}, further studies have analyzed the linguistic properties of synthetic data. \citet{liu2024benchmarking} explicitly focus on the ability of LLMs, Llama 3 and GPT-4o, to approximate child-caregiver dialogues. They test single-turn answer generation for child answers prompted with mother questions and vice versa, both in a zero- and few-shot setting. They find that in the single-turn setting (where the model produces one response and it's compared directly against a human reference utterance), models looked reasonably close to human data on the word-, utterance and dialogue level. In contrast, across multiple turns (where the model sustains a longer back-and-forth), the models' output diverged more sharply from human patterns, particularly on the utterance and discourse levels. In a similar vein, \citet{padovani2025dialogue} report that fine-tuning on natural dialogue continuations enables small Llama models to better approximate discourse coherence than fine-tuning on synthetic data.

\citet{ju2025domain} investigate on the syntactic properties of generated text, without focusing on interactive/developmental aspects. They find that when comparing human-written text to LLM-regenerated text (prompted with the beginning of the original human-written text), across the majority of their metrics (Flesch-Kincaid Scores, Dependency Tags, Depth Scores, Yngve Scores, Constituency Labels) and datasets (Wikipedia, CCNews and ELI5), models generate distributionally less diverse data, as evidenced by both a \textit{decrease in variability and a reduced long tail}.

Finally, \citet{nieth2026how} use corpus-linguistic measures in a context-aware evaluation framework, where human-likeness is assessed using a two-sample problem between the linguistic feature distribution of a human reference corpus for a given register and a corresponding LLM-generated corpus. Across all tested setups, LLMs deviate from the human baseline. Crucially, which models are closest to human language depends on the register, not model size: especially spoken data is hard to approximate with LMs, whereas synthetic WikiHow instructional texts are closer to human data.

\section{Data pre-processing} As our synthetic child-caregiver baseline, we use the aforementioned TinyDialogues (TD). We restrict our analysis to the evaluation split, as it already provides a sufficient amount of data (19k dialogues, 280k utterances). 
Individual conversations are parsed by splitting on included delimiters, turns are extracted via the included speaker labels. We construct comparable natural-language datasets by matching two properties of TD: a target dialogue length of approximately 201 words, averaged across the four synthetic age categories (2, 5, 10, and 15 years), and a target total word count of approximately 940,000 words (average over TD datasets).
To build the natural data, we collect child-caregiver dialogue from CHILDES \cite{macwhinney2000childes}, adult-adult interactions from TalkBank \cite{canavan1996callfriend,canavan1997callhome}, and classroom interactions from ClassBank \cite{macwhinney2014transcript}. 
All corpora are in CHAT format. We extract speaker-labeled utterances and clean CHAT-specific transcription conventions to match the structure of the synthetic files. Natural dialogues longer than the target segment length are cut at approximately 201 words, shorter dialogues are retained entirely. These segments are then shuffled and sampled until each comparison dataset reaches its total word budget of 940,000 words. For the classroom corpus, available data was insufficient to reach this budget, so we use all available data in full. For CHILDES data, we further separate between toddler (up to 3 years) and child (over 3 years). We expect the natural datasets to roughly correspond to the synthetic data in linguistic make-up (TD-2 to Toddler data, TD-5 to Child data, TD-10 to Classroom data and TD-15 to Adult interactions).

\begin{figure}[t]
\centering
\includegraphics[width=0.40\textwidth]{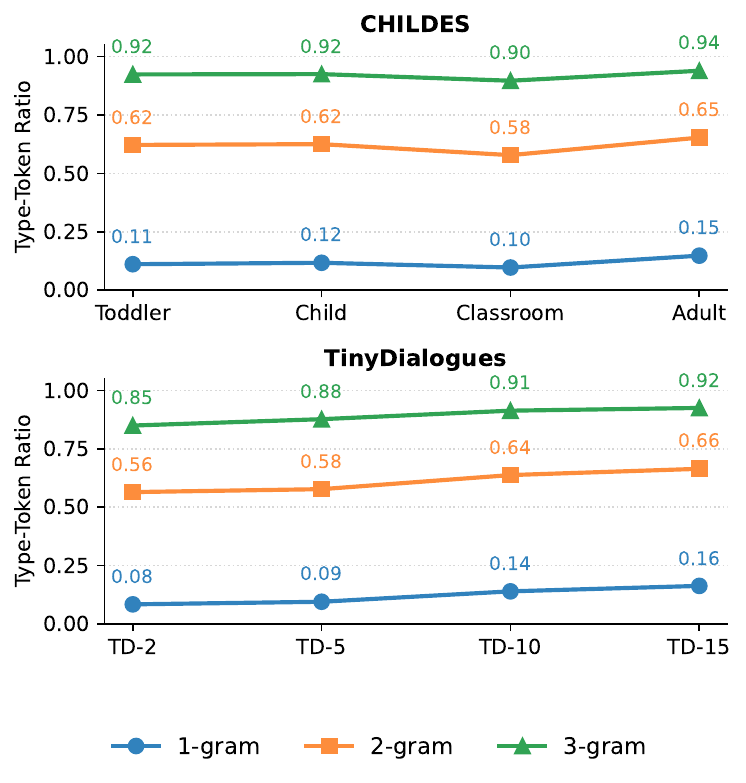}
\caption{Type-Token Ratio per dataset, computed over unigrams, bigrams, and trigrams.}
\label{fig:ttr}
\end{figure}

\section{Analyses}

We analyze natural/synthetic dialogues along several linguistic dimensions in the following sections.

\subsection{Type-token ratio}
To examine lexical diversity, we compute Type-token ratio (TTR) for each corpus at the unigram, bigram, and trigram level. For each n-gram size, we draw 10 random samples accumulating a fixed number of tokens (20,000) per sample, compute the TTR on the pooled n-grams, and average the resulting scores. 

\paragraph{Results} 
In general, Figure~\ref{fig:ttr} shows that TTR is relatively aligned between natural and synthetic data. Natural corpora are more lexically diverse for younger groups, but this difference vanishes as simulated age increases. However, it remains questionable if the gap at younger ages does reflect a genuinely higher lexical diversity in natural language. Especially in CHILDES, children's early production attempts, including mispronunciations and non-standard word forms, are sometimes pseudo-phonetically transcribed (e.g. \textit{wadda} instead of \textit{water}), inflating the count of unique forms. Within the natural data, Classroom has the lowest TTR, likely reflecting the narrow thematic scope, while Adult dialogue shows the highest lexical variety, consistent with an expected broader vocabulary. Synthetic TTR grows steadily, with TD-10 and TD-15 converging towards the natural Adult data. This developmental trajectory suggests that the LLM generates increasingly varied vocabulary as the simulated interlocutor ages grow, and that the reduced lexical variability reported for LLM-generated text \citep{ju2025domain} is modulated by the targeted developmental stage.

\begin{figure}[t]
\centering
\includegraphics[width=0.4\textwidth]{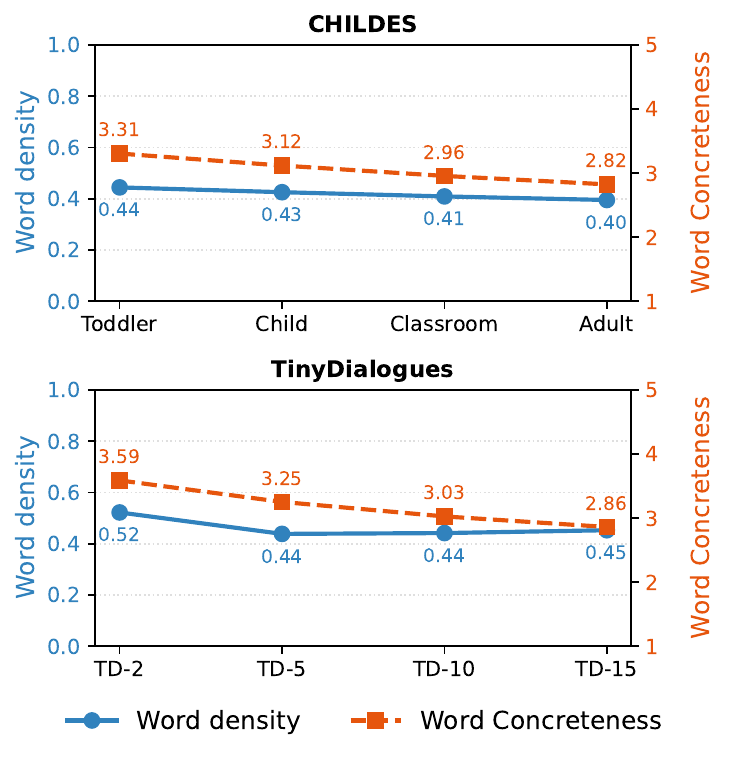}
\caption{Distribution of word density and word concreteness per dataset.}
\label{fig:density_concreteness}
\end{figure}

\subsection{Word density and concreteness}
Following \citet{liu2024benchmarking}, we compute two further word-level measures. \textbf{Word density} is defined as the proportion of content words (nouns, proper nouns, verbs, adjectives, and adverbs) relative to all other words in an utterance, 
and represents the informational load of an utterance. Following \citet{dawson2021features}, it is expected to decrease with age, as speech shifts from telegraphic to more function-word-rich. \textbf{Word concreteness} is computed as the mean concreteness rating of content-word lemmas in an utterance, using the human concreteness norms of \citet{brysbaert2014concreteness} (rated on a 1–5 scale, from abstract to concrete), averaged across utterances with at least one rated content word. Concreteness is expected to decrease with age as well, as children's early speech centers on concrete, here-and-now objects and actions, while older children and adults increasingly discuss more abstract, social, and epistemic topics. 

\paragraph{Results} Figure~\ref{fig:density_concreteness} shows close alignment again. Word density decreases with age in both data types, reflecting the shift from telegraphic to more function-word-rich speech. The clearest departure from natural data is at the youngest synthetic age, where TD-2 (0.52) is notably higher than Toddler (0.44), suggesting overly telegraphic toddler speech. Older synthetic ages converge toward natural values. Concreteness follows the same developmental arc in both data types, decreasing with age as conversations shifts from concrete to more abstract topics. Synthetic data is consistently, though only slightly, higher in concreteness than natural data at every age, while closely tracking the same downward trend.

\begin{table}[t]
\centering
\footnotesize
\sisetup{
  detect-weight = true,
  detect-family = true
}
\begin{tabular}{l
                S[table-format=6.0]
                S[table-format=2.1]
                c}
\toprule
\textbf{Dataset} & {\makecell{\textbf{Total}\\\textbf{Utterances}}} & {\makecell{\textbf{Avg.}\\\textbf{Turns}}} & {\makecell{\textbf{Min/Max}\\\textbf{Turns}}} \\
\midrule
Toddler   & 273375 & \bfseries 57.5 & 10/128 \\
Child     & 216924 & 45.3          & 9/128  \\
Classroom & 37704  & 26.1          & 2/113  \\
Adult     & 78505  & 31.6          & 3/68   \\

\midrule
TD-2   & 103188 & \bfseries 15.7 & 4/31  \\
TD-5   & 69215  & 13.7          & 1/28   \\
TD-10  & 56077  & 13.5          & 3/28   \\
TD-15  & 49951  & 12.8          & 2/28   \\
\bottomrule
\end{tabular}
\caption{Dialogue structure per dataset.}
\label{tab:dialogue_structure}
\end{table}

\subsection{Average utterance length and average turns}
For each dialogue, we count the number of conversational turns and measure the average utterance length, further separating by speaker role (child CHI vs.\ caregiver CAR). We then compute, for each dataset (age-ordered TD, natural data), the average number of turns, the average utterance length and minimum/maximum number of turns.

\paragraph{Results}

Table~\ref{tab:dialogue_structure} summarizes dialogue-level statistics. Since natural and synthetic data are matched on total word count, turn count and utterance length necessarily trade off against each other. Figure~\ref{fig:length_depth} (left) illustrates that within each data type, overall utterance length increases with age (Toddler < Child < Adult < Classroom for natural data; TD-2 < TD-5 < TD-10 < TD-15 for synthetic data). Synthetic utterances are disproportionately longer than natural ones at every stage, most strikingly at TD-2, whose overall average (8.2 tokens) already exceeds natural Classroom and Adult. This inflated average is not evenly shared between speakers, at least at the youngest synthetic age. In natural data, caretaker (CAR) utterances are only modestly longer than child (CHI) utterances throughout, and this asymmetry stays fairly stable with age, even reversing slightly in Classroom. In synthetic data, the asymmetry is far wider at TD-2, where the high overall average is driven almost entirely by excessively long caregiver turns, while synthetic children speak in utterances shorter than real toddlers. From TD-5 onward, however, both CHI and CAR utterances are far longer than anything observed in natural data at every age, with even the shorter of the two speaker roles well exceeding the longest natural category. In other words, only TD-2 owes its inflated average primarily to the caregiver role. For all older synthetic ages, both speakers are simply generated with excessively long turns, indicating that TinyDialogues systematically overshoots natural utterance length.

\subsection{Average parse tree depth}
We compute average dependency tree depth, defined as the length of the longest path from the root to any leaf node. To do that, we add UD annotations to all datasets with the CAIT toolkit \cite{padovani-etal-2026-cait}, which combines a dependency parser and a POS-tagger trained on CHILDES data. We average tree depth across all utterances in each file, further broken down by speaker role (CHI vs.\ CAR).

\begin{figure}[t]
\centering
\includegraphics[width=0.50\textwidth]{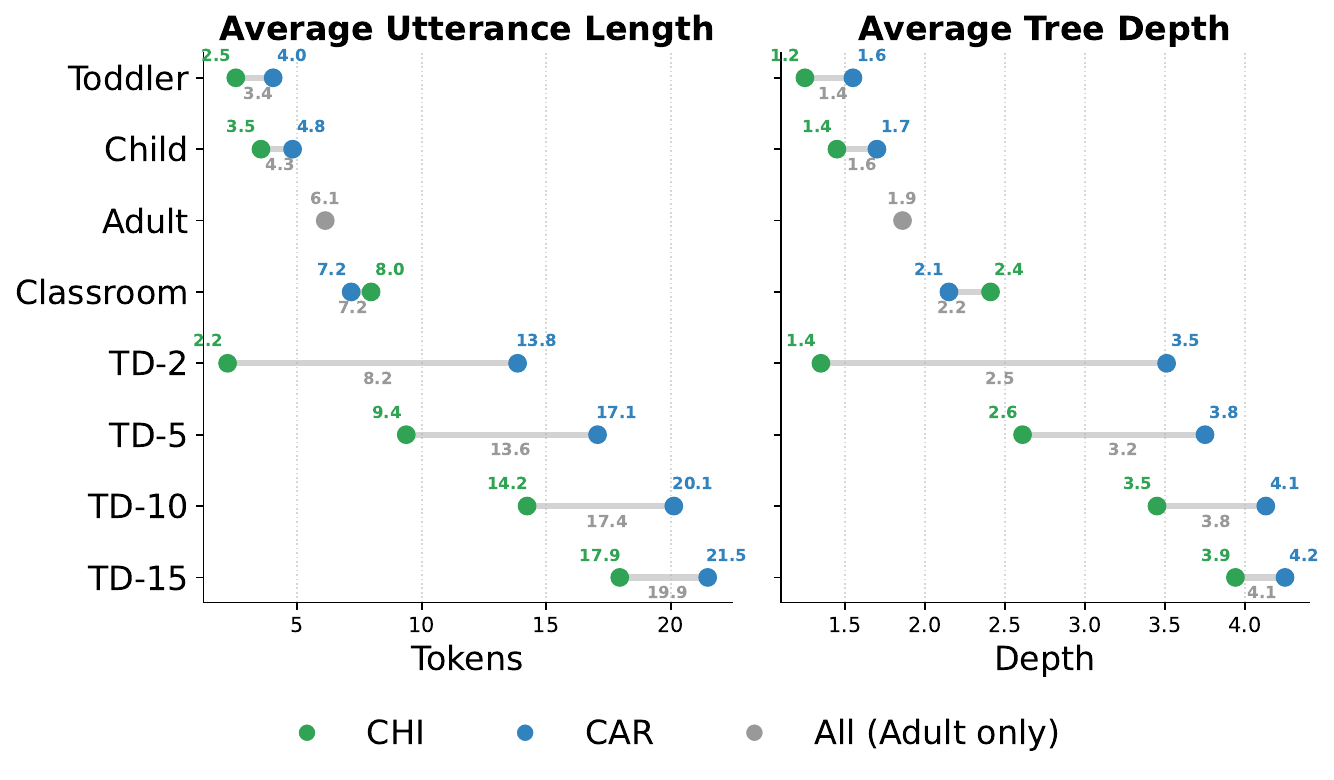}
\caption{Average utterance length (tokens) and average dependency tree depth per dataset, overall and broken down by speaker role (CHI vs.\ CAR).}
\label{fig:length_depth}
\end{figure}

\paragraph{Results} Average tree depth, depicted in Figure~\ref{fig:length_depth} (right), follows the same trend as utterance length. Overall depth increases with age in both data types and is consistently higher for synthetic dialogues at every stage. This is unsurprising, given that longer utterances generally yield deeper syntactic structures. Breaking this down by speaker role reveals the same asymmetry as before. At TD-2, the inflated depth is again driven almost entirely by the caregiver role (3.5), with child utterances (1.4) close to Toddler depth. From TD-5 onward, this excess complexity is no longer specific to caregiver speech, but affects both speaker roles.

\subsection{Utterance-level constructions}
To complement average parse tree depth with a more interpretable view of syntactic complexity, we use the construction tagger integrated in the CAIT toolkit \citep{padovani-etal-2026-cait}, which classifies utterances into a fixed set of construction types \cite{cameron2003construction,bunzeck2025richness}: formulaic (FOR), fragments without a predicate (FRA), wh- or yes/no-questions (QWH, QYN), copula constructions (COP), imperatives (IMP), subject-predicate in/transitive constructions (SPI, SPT), and complex utterances with at least two predicates (COM). 

\begin{figure}[t]
\centering
\includegraphics[width=0.4\textwidth]{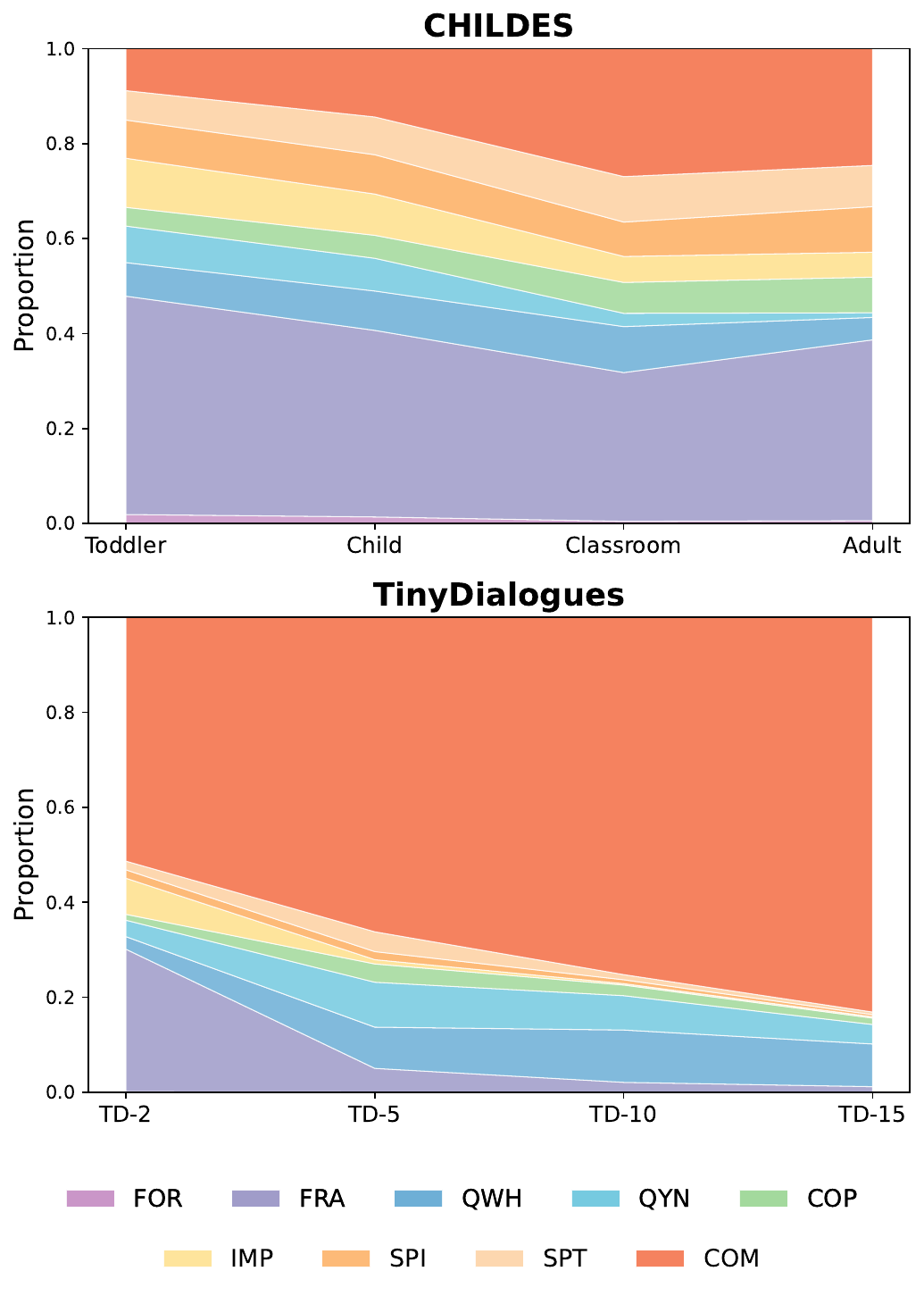}
\caption{Distribution of utterance-level construction types (\%) per dataset, as tagged by the CAIT construction tagger.}
\label{fig:constructions}
\end{figure}

\paragraph{Results} Figure \ref{fig:constructions} show the distribution and development of these construction types. The most striking contrasts emerge for fragments and complex utterances. Natural corpora show high fragment rates (31.31–46.07\%), reflecting characteristics of spoken data such as backchannels, hesitations, repairs, or truncated turns. Synthetic data drops to near-zero fragment rates (0.01–4.97\%), except for TD-2 (29.99\%), which partially captures the telegraphic speech typical of infants. Conversely, complex utterances dominate synthetic corpora (51.34–83.12\%), but are far less prevalent in natural speech (9.02–26.84\%). This result is consistent with the higher tree depths observed above for synthetic data. Another  divergence can be found for questions (QWH and QYN), which become less frequent in natural data (consistent with findings by \citealp{bunzeck2025richness}), but actually more frequent in synthetic data.

\subsection{Semantic alignment}
\label{sec:semantic_alignment}
To characterize how semantically aligned speakers are within a dialogue, we compute two complementary measures, both relying on a sentence-transformer model \texttt{sentence-transformers/all-MiniLM-L6-v2} that maps utterances to a 384-dimensional dense vector space \cite{reimers2019sentence}. As a global measure, \textbf{Inter-Speaker Similarity} captures whether speakers remain on a shared topic across the conversation. For each dialogue, utterances are grouped by speaker, encoded, and averaged to single vector per speaker. We then calculate the cosine similarity between every pair of speakers and average across dialogues to obtain one score per dataset. This measure is insensitive to turn order. As a local measure, \textbf{Turn-by-turn Alignment}, following \citet{guo2024curious}, captures whether each utterance semantically tracks the utterance that immediately precedes it, based on adjacent turn pairs. For each pair of consecutive turns from different speakers, we compute the cosine similarity between the two utterance embeddings, average within each dialogue, and average across dialogues to obtain one score per dataset. Unlike the previous measure, this metric is sensitive to conversational order and captures local, turn-by-turn semantic contingency.

\paragraph{Results} Both metrics are reported in Table~\ref{tab:semantic_alignment}. Inter-Speaker Similarity declines with age in both natural and synthetic data, following a broadly similar trajectory. In both types of data, similarity is highest for young age groups, where talk revolves around concrete objects in the immediate environment, marked by frequent repetition to reinforce concepts and seek approval from interlocutors. For older age-groups, inter-speaker similarity decreases.
Turn-by-turn Alignment also decreases with age in both data types. Adjacent turns are most semantically aligned for the youngest speakers, following the same directional trend in both natural and synthetic data. Synthetic dialogues, however, consistently show much higher scores than natural ones at every age, indicating that automatically generated conversations remain considerably more coherent turn-by-turn, whereas natural data diverges more strongly even between individual turns.

\begin{table}[t]
\centering
\scriptsize
\sisetup{
  detect-weight = true,
  detect-family = true
}
\begin{tabular}{l
                S[table-format=1.2]
                S[table-format=1.2]}
\toprule
\textbf{Dataset} & {\makecell{\textbf{Inter-Speaker}\\\textbf{Similarity}}} & {\makecell{\textbf{Turn-by-turn}\\\textbf{Alignment}}} \\
\midrule
Toddler   & \bfseries 0.75 & \bfseries 0.30 \\
Child     & 0.74          & 0.26 \\
Classroom & 0.52          & 0.27 \\
Adult     & 0.63          & 0.22 \\
\midrule
TD-2   & \bfseries 0.72 & \bfseries 0.49 \\
TD-5   & \bfseries 0.72 & 0.46 \\
TD-10  & 0.69          & 0.41 \\
TD-15  & 0.66          & 0.38 \\
\bottomrule
\end{tabular}
\caption{Inter-speaker semantic similarity and turn-by-turn semantic alignment per dataset.}
\label{tab:semantic_alignment}
\end{table}

\subsection{Linguistic alignment}
While Section~\ref{sec:semantic_alignment} examines whether speakers converge on the same meaning across a dialogue, alignment can also be assessed at the lexical and syntactic levels. We use the ALIGN library \citep{duran2019align} to compute lexical and syntactic alignment between adjacent speaker turns. For each dataset, we compute alignment between adjacent turns, average these turn-level scores within each dialogue, and then average across dialogues to obtain one score per dataset. Syntactic alignment reflects the degree to which a turn repeats the part-of-speech sequences of the immediately preceding turn, while lexical alignment reflects the degree to which it repeats the same word forms. We report alignment scores computed on lemmas at unigram and bigram level.

\begin{table}[t]
\centering
\scriptsize
\begin{tabular}{lrrrr}
\toprule
\textbf{Dataset} & \makecell{\textbf{Syntax}\\\textbf{(lemma,} \\\textbf{1-gram)}} & \makecell{\textbf{Syntax}\\\textbf{(lemma,} \\\textbf{2-gram)}} & \makecell{\textbf{Lexical}\\\textbf{(lemma,} \\\textbf{1-gram)}} & \makecell{\textbf{Lexical}\\\textbf{(lemma,} \\\textbf{2-gram)}} \\
\midrule
Toddler  & 0.32 & 0.10 & 0.24 & 0.09  \\
Child    & 0.36 & 0.11 & 0.22 & 0.07  \\
Classroom         & 0.41 & 0.13 & 0.23 & 0.07  \\
Adult             & 0.42 & 0.12 & 0.19 & 0.04  \\
\midrule
TD-2           & 0.38 & 0.11 & 0.21 & 0.04  \\
TD-5           & 0.51 & 0.16 & 0.22 & 0.04  \\
TD-10          & 0.59 & 0.20 & 0.23 & 0.03  \\
TD-15          & 0.65 & 0.24 & 0.23 & 0.02  \\
\bottomrule
\end{tabular}
\caption{Syntactic and lexical alignment (lemma-level, unigram and bigram) per dataset.}
\label{tab:alignment}
\end{table}

\paragraph{Results} Both values are reported in Table~\ref{tab:alignment}. In natural data, lexical alignment decreases with age, consistent with the decline already observed for turn-by-turn Similarity. In synthetic data, however, lexical alignment stays stable across age groups, slightly at odds with the declining trend captured by turn-by-turn Similarity. This is surprising, but might be caused by the longer utterances which are more complex and therefore allow for more alignment, while simultaneously spanning a broader semantic space.
Syntactic alignment, in contrast, increases steadily with age in both data types, though it remains consistently higher in synthetic dialogue throughout. While it is generally not surprising that children might mirror their input structurally when acquiring grammar, the synthetic data stays unnaturally repetitive on the structural level.

\subsection{Children's grammatical errors}

\begin{figure*}[t]
\centering
\includegraphics[width=\textwidth]{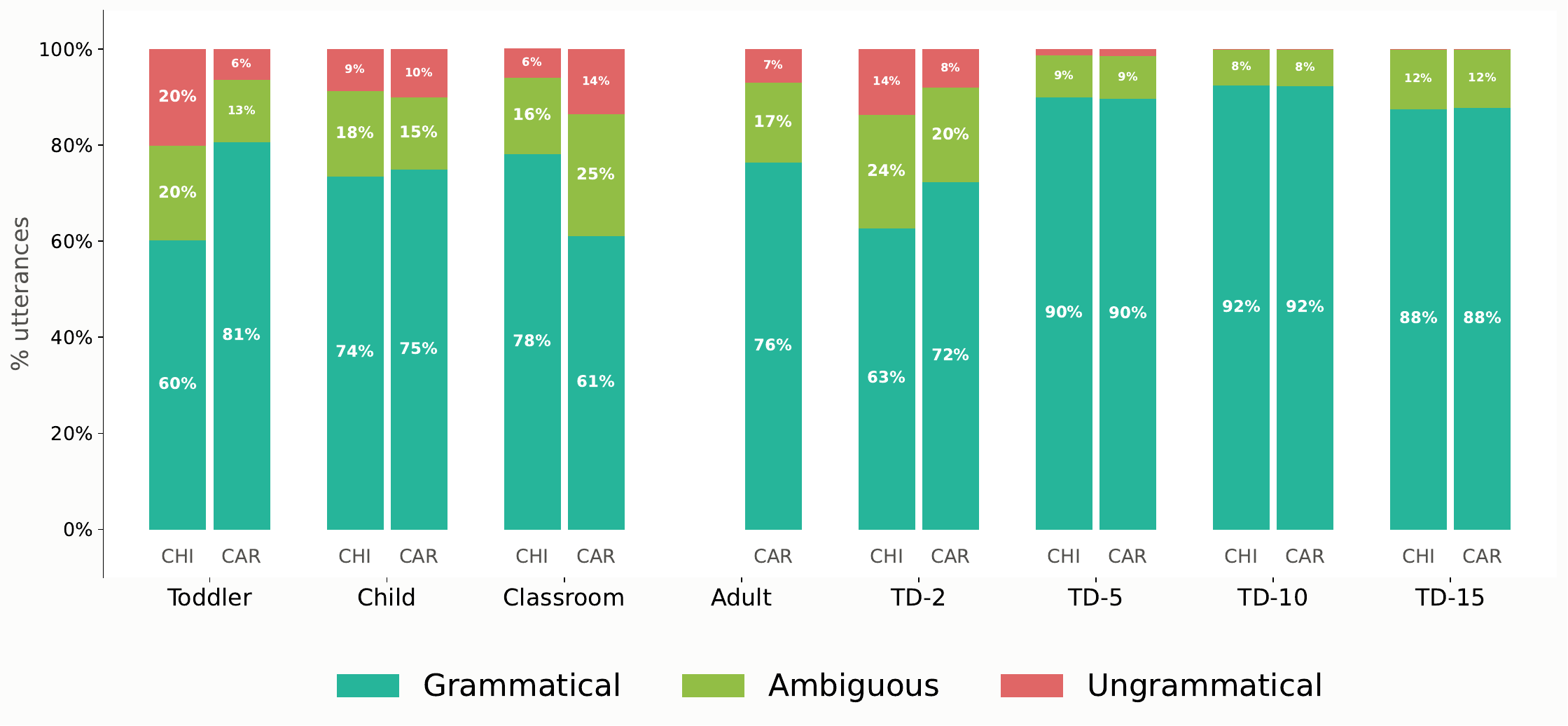}
\caption{Proportion of grammatical, ambiguous, and ungrammatical utterances per dataset, split by speaker role (CHI vs.\ CAR).}
\vspace{-0.5em}
\label{fig:grammaticality}
\end{figure*}

Finally, we assess the ``grammaticality'' of utterances using the context-dependent classifier of \citet{nikolaus2024automatic}, trained to annotate child-caregiver conversations from CHILDES. As the authors' model checkpoints are not publicly available, we retrain the classifier from scratch on their released annotated data and use the best checkpoint for prediction. We map all speakers across our natural and synthetic datasets onto a binary speaker variable, CHI and CAR, the latter encompassing all conversational partners. For each dataset, we compute the proportion of grammatical, ungrammatical, and ambiguous sentences predicted by the model, reported in Figure~\ref{fig:grammaticality}. 

\paragraph{Results} Figure \ref{fig:grammaticality} shows that child speech is least grammatical at the youngest age in both natural and synthetic data, and that grammaticality rises as children grow older. In the natural data, grammaticality increases only modestly, and a substantial share of ungrammatical or ambiguous utterances persists even in the older age groups. In the synthetic data, by contrast, grammaticality rises sharply by age 5, peaks at age 10, and drops only slightly at age 15. For older age groups, ungrammatical utterances become rare, while ambiguous ones become comparatively more frequent. This may indicate that the complexity of utterances produced by older speakers is not  captured by our classifier, which was trained on CHILDES data drawn primarily from younger children, mainly between 2 and 5 years of age. Caregiver speech follows a broadly similar developmental pattern to child speech once children are above 2 years old, with one notable exception: in Classroom, CAR grammaticality (61\%) falls markedly below CHI (78\%). When children are younger, caregiver grammaticality is consistently higher than child grammaticality, though still containing a substantial portion of grammatical mistakes, likely due to the simplified, telegraphic nature of child-directed speech, which is annotated as ungrammatical.

\section{Fine-tuning approach}
\begin{figure*}[t]
\centering
\includegraphics[width=0.8\textwidth]{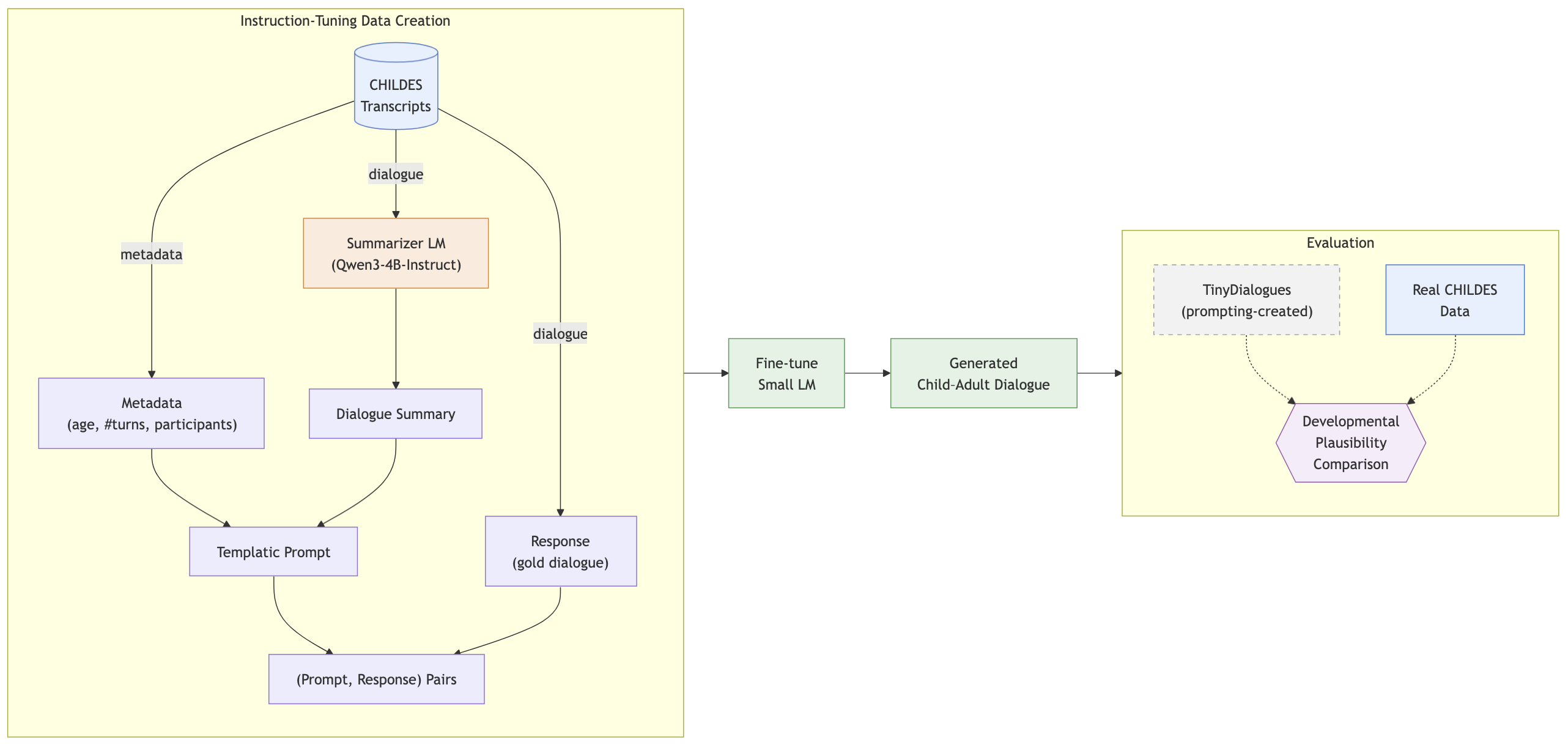}
\caption{Fine-tuning workflow from natural data.}
\label{fig:finetuning}
\end{figure*}

As we have shown that natural and synthetic data diverge on a variety of linguistic measures, we propose a novel fine-tuning approach to generate more developmentally plausible dialogue. To do so, we use instruction-tuning data generated from real dialogues. We create a novel instruction-tuning dataset with prompts that are templatically created from CHILDES metadata (child age, number of turns, participants) and an LM-generated dialogue summary. The original transcript is used as the gold response (cf. Figure~\ref{fig:finetuning}). As of now, we can only report preliminary results, but they are nonetheless promising. We used \texttt{unsloth} to fine-tune \texttt{qwen3-4b-instruct} on roughly 5\% of the data that we have created from CHILDES. We then used this model with novel, hand-written summary prompts inserted into the template used for fine-tuning to generate 250 dialogues (max. 300 novel tokens each) from 25 hand-written prompts (see Appendix \ref{sec:prompts}). We used nucleus sampling with a temperature of 1.2, as lower temperature settings lead to overly repetitive dialogue. While repetition is normal and expected in child-caretaker dialogues, some of the generations at temperature 0.8 consisted of sole single-word repetitions. 

For space reasons, we only present one evaluation of this novel dialogue generated with our small, 4B-parameter model. For utterance-level construction analyses, it actually approximates the construction-level statistics of natural data after a moderate amount of parameter-efficient fine-tuning (Table~\ref{tab:cxn-ft}): unlike TD, it reproduces natural rates of fragments, imperatives and subject-predicate constructions, without the synthetic overshoot in complex utterances reported earlier. This suggests that fine-tuning on real interaction data is a viable path towards developmentally plausible synthetic dialogue.

\begin{table*}[h!]
\centering
{\scriptsize
\renewcommand{\arraystretch}{1.0}
\begin{tabular}{lrrrrrrrrr}
\toprule
\textbf{Dataset} & \textbf{FOR} & \textbf{FRA} & \textbf{QWH} & \textbf{QYN} & \textbf{COP} & \textbf{IMP} & \textbf{SPI} & \textbf{SPT} & \textbf{COM} \\
\midrule
Toddler & 1.9 & 46.0 & 7.1 & 7.7 & 4.0 & 10.3 & 8.1 & 6.2 & 8.9 \\
Child & 1.4 & 39.3 & 8.3 & 6.9 & 4.8 & 8.7 & 8.3 & 8.0 & 14.4 \\
Classroom & 0.5 & 31.3 & 9.7 & 2.8 & 6.5 & 5.5 & 7.2 & 9.6 & 27.0 \\
Adult & 0.6 & 38.1 & 4.7 & 1.0 & 7.5 & 5.3 & 9.6 & 8.7 & 24.6 \\
\midrule
TD-2 & 0.1 & 30.0 & 2.7 & 3.5 & 1.3 & 7.6 & 1.8 & 1.8 & 51.3 \\
TD-5 & 0.1 & 5.0 & 8.7 & 9.5 & 3.9 & 0.9 & 1.7 & 4.2 & 66.2 \\
TD-10 & 0.0 & 2.1 & 11.0 & 7.2 & 2.2 & 0.3 & 0.8 & 1.1 & 75.2 \\
TD-15 & 0.0 & 1.2 & 9.0 & 4.1 & 1.4 & 0.3 & 0.5 & 0.5 & 83.1 \\
\midrule
Fine-tuned \texttt{Qwen3-4B-Instruct} (ours) & 0.9 & 39.4 & 8.0 & 4.5 & 5.0 & 10.9 & 10.7 & 11.9 & 8.6 \\
\bottomrule
\end{tabular}
}
\caption{Distribution of utterance-level construction types (\%) per dataset: formulaics (FOR), fragments (FRA), wh-/yes-no questions (QWH, QYN), copula (COP), imperatives (IMP), subject-predicate in/transitives (SPI, SPT), complex utterances (COM).}
\vspace{-1em}
\label{tab:cxn-ft}
\end{table*}

\section{Discussion and conclusion}

Overall, our results show that synthetic dialogue diverges from natural dialogue on key linguistic measures: it consists of overly verbose, complex sentences and very long utterances, is highly grammatical and syntactically more complex, yet lacks construction-level diversity, and is overly aligned on the semantic level. A plausible failure mode underlying these patterns is that TinyDialogues does not become more plausible as the simulated child grows older — it merely drifts toward the LLM's home register of written, complete-sentence text, as evidenced by TD-10/15 converging on natural adult lexical diversity while overshooting every natural category in utterance length, tree depth, and complex constructions. What is missing is speech itself: fragments, repairs, backchannels, and errors are not noise but constitutive of spoken interaction. Their absence could also explain the benchmark successes of TD-trained models \cite{feng2024childdirected}, which may profit from unnaturally complete and complex input rather than from a better simulation of developmental data. These findings align with previous studies on the uniformity of synthetic text \cite{ju2025domain,liu2024benchmarking,nieth2026how} and with \citet{bunzeck2026child-directed}, who report that TinyDialogues-trained models, unlike CHILDES-trained ones, fail to predict discourse-deictic expressions central to acquisition. Taken together, it is questionable whether such synthetic data should be used in developmentally plausible modeling studies.

Our pilot study, however, suggests that this might be a data problem, not a general lack of capacity: already a moderate amount of parameter-efficient fine-tuning on real transcripts brings a small 4B model much closer to natural construction distributions than naive prompting of a much larger model. This echoes recent work that fine-tunes on human data to capture human behavior \cite{binz2025foundation}. Still, these results are preliminary: Whether the generated dialogues are also semantically coherent, faithfully age-conditioned, and ultimately useful as training data for developmentally plausible LMs remains open.





\paragraph{Future work}
We identify multiple avenues for future work. Our analyses could be extended a) on the linguistic level (including even more corpus-linguistic measures), b) on the language level (expanding to synthetic data in other languages, such as the code-switched data by \citealp{zeng2026bringing}), and c) on the genre level, e.g. to simple synthetic stories \cite{eldan2023tinystories,finke2025parameterized} and their natural counterparts. All these approaches would shed even more light on the linguistic dimensions of LM-generated language and their human-likeness.

\bibliography{custom}

@inproceedings{bunzeck2026child-directed,
    title = "Child-directed speech facilitates production, not comprehension, in {B}aby{LM}s",
    author = "Bunzeck, Bastian  and
      Zarrie{\ss}, Sina",
    editor = "Bonial, Claire  and
      Berzak, Yevgeni",
    booktitle = "Proceedings of the 30th Conference on Computational Natural Language Learning",
    month = jul,
    year = "2026",
    address = "San Diego, California, USA",
    publisher = "Association for Computational Linguistics",
    url = "https://aclanthology.org/2026.conll-main.14/",
    doi = "10.18653/v1/2026.conll-main.14",
    pages = "227--249",
    ISBN = "979-8-89176-410-1"
}

@misc{canavan1997callhome,
  author       = {Canavan, Alexandra and Graff, David and Zipperlen, George},
  title        = {{CALLHOME American English Speech LDC97S42}},
  year         = {1997},
  publisher    = {Linguistic Data Consortium},
  address      = {Philadelphia, PA},
}

@misc{canavan1996callfriend,
  author       = {Canavan, Alexandra and Zipperlen, George},
  title        = {{CALLFRIEND American English--Non-Southern Dialect LDC96S46}},
  year         = {1996},
  publisher    = {Linguistic Data Consortium},
  address      = {Philadelphia, PA},
}

@article{binz2025foundation,
  title = {A Foundation Model to Predict and Capture Human Cognition},
  author = {Binz, Marcel and Akata, Elif and Bethge, Matthias and Br{\"a}ndle, Franziska and Callaway, Fred and {Coda-Forno}, Julian and Dayan, Peter and Demircan, Can and Eckstein, Maria K. and {\'E}ltet{\H o}, No{\'e}mi and Griffiths, Thomas L. and Haridi, Susanne and Jagadish, Akshay K. and {Ji-An}, Li and Kipnis, Alexander and Kumar, Sreejan and Ludwig, Tobias and Mathony, Marvin and Mattar, Marcelo and Modirshanechi, Alireza and Nath, Surabhi S. and Peterson, Joshua C. and Rmus, Milena and Russek, Evan M. and Saanum, Tankred and Schubert, Johannes A. and Schulze Buschoff, Luca M. and Singhi, Nishad and Sui, Xin and Thalmann, Mirko and Theis, Fabian J. and Truong, Vuong and Udandarao, Vishaal and Voudouris, Konstantinos and Wilson, Robert and Witte, Kristin and Wu, Shuchen and Wulff, Dirk U. and Xiong, Huadong and Schulz, Eric},
  year = 2025,
  month = aug,
  journal = {Nature},
  volume = {644},
  number = {8078},
  pages = {1002--1009},
  issn = {0028-0836, 1476-4687},
  doi = {10.1038/s41586-025-09215-4},
  url = {https://www.nature.com/articles/s41586-025-09215-4},
  urldate = {2026-01-11},
  langid = {english}
}

@misc{eldan2023tinystories,
  title = {{{TinyStories}}: {{How Small Can Language Models Be}} and {{Still Speak Coherent English}}?},
  shorttitle = {{{TinyStories}}},
  author = {Eldan, Ronen and Li, Yuanzhi},
  year = 2023,
  month = may,
  number = {arXiv:2305.07759},
  eprint = {2305.07759},
  primaryclass = {cs},
  publisher = {arXiv},
  url = {http://arxiv.org/abs/2305.07759},
  urldate = {2023-07-04},
  archiveprefix = {arXiv}
}

@inproceedings{feng2024childdirected,
  title = {Is Child-Directed Speech Effective Training Data for Language Models?},
  booktitle = {Proceedings of the 2024 Conference on Empirical Methods in Natural Language Processing},
  author = {Feng, Steven Y. and Goodman, Noah and Frank, Michael},
  editor = {{Al-Onaizan}, Yaser and Bansal, Mohit and Chen, Yun-Nung},
  year = 2024,
  month = nov,
  pages = {22055--22071},
  publisher = {Association for Computational Linguistics},
  address = {Miami, Florida, USA},
  url = {https://aclanthology.org/2024.emnlp-main.1231}
}

@inproceedings{finke2025parameterized,
  title = {Parameterized Synthetic Text Generation with {{SimpleStories}}},
  booktitle = {The Thirty-Ninth Annual Conference on Neural Information Processing Systems Datasets and Benchmarks Track},
  author = {Finke, Lennart and Sreedhara, Chandan and Dooms, Thomas and Allen, Mat and Rodriguez, Juan Diego and Nabeshima, Noa and Marshall, Thomas and Braun, Dan},
  year = 2025,
  url = {https://openreview.net/forum?id=sVh3eQ642W}
}

@inproceedings{haga2024babylm,
  title = {{{BabyLM Challenge}}: {{Exploring}} the Effect of Variation Sets on Language Model Training Efficiency},
  booktitle = {The 2nd {{BabyLM}} Challenge at the 28th Conference on Computational Natural Language Learning},
  author = {Haga, Akari and Fukatsu, Akiyo and Oba, Miyu and Bisazza, Arianna and Oseki, Yohei},
  editor = {Hu, Michael Y. and Mueller, Aaron and Ross, Candace and Williams, Adina and Linzen, Tal and Zhuang, Chengxu and Choshen, Leshem and Cotterell, Ryan and Warstadt, Alex and Wilcox, Ethan Gotlieb},
  year = 2024,
  month = nov,
  pages = {252--261},
  publisher = {Association for Computational Linguistics},
  address = {Miami, FL, USA},
  url = {https://aclanthology.org/2024.conll-babylm.23/}
}

@inproceedings{ju2025domain,
  title = {Domain {{Regeneration}}: {{How}} Well Do {{LLMs}} Match Syntactic Properties of Text Domains?},
  booktitle = {Findings of the Association for Computational Linguistics: {{ACL}} 2025},
  author = {Ju, Da and Blix, Hagen and Williams, Adina},
  editor = {Che, Wanxiang and Nabende, Joyce and Shutova, Ekaterina and Pilehvar, Mohammad Taher},
  year = 2025,
  month = jul,
  pages = {2367--2388},
  publisher = {Association for Computational Linguistics},
  address = {Vienna, Austria},
  url = {https://aclanthology.org/2025.findings-acl.120/},
  isbn = {979-8-89176-256-5}
}

@inproceedings{lee2025readability,
  title = {Readability {$\neq$} Learnability: {{Rethinking}} the Role of Simplicity in Training Small Language Models},
  booktitle = {Second Conference on Language Modeling},
  author = {Lee, Ivan and {Berg-Kirkpatrick}, Taylor},
  year = 2025,
  url = {https://openreview.net/forum?id=AFMGbq39bQ}
}

@misc{liu2024benchmarking,
  title = {Benchmarking {{LLMs}} for {{Mimicking Child-Caregiver Language}} in {{Interaction}}},
  author = {Liu, Jing and Fourtassi, Abdellah},
  year = 2024,
  publisher = {arXiv},
  doi = {10.48550/ARXIV.2412.09318},
  url = {https://arxiv.org/abs/2412.09318},
  urldate = {2025-06-27},
  copyright = {Creative Commons Attribution 4.0 International}
}

@book{macwhinney2000childes,
  title = {The {{CHILDES}} Project: {{Tools}} for Analyzing Talk},
  author = {MacWhinney, Brian},
  year = 2000,
  edition = {3},
  publisher = {Lawrence Erlbaum Associates},
  address = {Mahwah, NJ}
}

@misc{nieth2026how,
  title = {How {{Human-Like Are Large Language Models}}? {{A Register-Aware Linguistic Evaluation Framework}}},
  shorttitle = {How {{Human-Like Are Large Language Models}}?},
  author = {Nieth, Bj{\"o}rn and Gracheva, Marianna and Mahlberg, Michaela and Eskofier, Bjoern and Salin, Emmanuelle},
  year = 2026,
  month = may,
  number = {arXiv:2605.23651},
  eprint = {2605.23651},
  primaryclass = {cs.CL},
  publisher = {arXiv},
  doi = {10.48550/arXiv.2605.23651},
  url = {http://arxiv.org/abs/2605.23651},
  urldate = {2026-05-27},
  archiveprefix = {arXiv}
}

@inproceedings{nikolaus2024automatic,
  title = {Automatic Annotation of Grammaticality in Child-Caregiver Conversations},
  booktitle = {Proceedings of the 2024 Joint International Conference on Computational Linguistics, Language Resources and Evaluation ({{LREC-COLING}} 2024)},
  author = {Nikolaus, Mitja and Agrawal, Abhishek and Kaklamanis, Petros and Warstadt, Alex and Fourtassi, Abdellah},
  editor = {Calzolari, Nicoletta and Kan, Min-Yen and Hoste, Veronique and Lenci, Alessandro and Sakti, Sakriani and Xue, Nianwen},
  year = 2024,
  month = may,
  pages = {1832--1844},
  publisher = {{ELRA and ICCL}},
  address = {Torino, Italia},
  url = {https://aclanthology.org/2024.lrec-main.164/}
}

@inproceedings{padovani2025dialogue,
  title = {Dialogue Is Not Enough to Make a Communicative {{BabyLM}} (but Neither Is Developmentally Inspired Reinforcement Learning)},
  booktitle = {Proceedings of the First {{BabyLM}} Workshop},
  author = {Padovani, Francesca and Bunzeck, Bastian and Ali, Manar and Momen, Omar and Bisazza, Arianna and Buschmeier, Hendrik and Zarrie{\ss}, Sina},
  editor = {Charpentier, Lucas and Choshen, Leshem and Cotterell, Ryan and Gul, Mustafa Omer and Hu, Michael Y. and Liu, Jing and Jumelet, Jaap and Linzen, Tal and Mueller, Aaron and Ross, Candace and Shah, Raj Sanjay and Warstadt, Alex and Wilcox, Ethan Gotlieb and Williams, Adina},
  year = 2025,
  month = nov,
  pages = {421--435},
  publisher = {Association for Computational Linguistics},
  address = {Suzhou, China},
  url = {https://aclanthology.org/2025.babylm-main.29/},
  isbn = {TODO}
}

@inproceedings{rasanen2024agedependent,
  title = {Age-Dependent Analysis and Stochastic Generation of Child-Directed Speech},
  booktitle = {Proceedings of the {{Annual Meeting}} of the {{Cognitive Science Society}}},
  author = {R{\"a}s{\"a}nen, Okko and Kocharov, Daniil},
  year = 2024,
  volume = {46},
  url = {https://escholarship.org/uc/item/1gz1458q}
}

@misc{zeng2026bringing,
  title = {Bringing {{Up}} a {{Bilingual BabyLM}}: {{Investigating Multilingual Language Acquisition Using Small-Scale Models}}},
  shorttitle = {Bringing {{Up}} a {{Bilingual BabyLM}}},
  author = {Zeng, Linda and Feng, Steven Y. and Frank, Michael C.},
  year = 2026,
  publisher = {arXiv},
  doi = {10.48550/ARXIV.2603.29552},
  url = {https://arxiv.org/abs/2603.29552},
  urldate = {2026-05-27},
  copyright = {Creative Commons Attribution 4.0 International}
}

@incollection{macwhinney2014transcript,
  title={A transcript-video database for collaborative commentary in the learning sciences},
  author={MacWhinney, Brian},
  booktitle={Video research in the learning sciences},
  pages={537--546},
  year={2014},
  publisher={Routledge}
}

@inproceedings{padovani-etal-2026-cait,
    title = "{CAIT}: A Syntactic Parsing Toolkit for Child{--}Adult {I}n{T}eractions",
    author = "Padovani, Francesca  and
      Yang, Xiulin  and
      Bunzeck, Bastian  and
      Jumelet, Jaap  and
      Matusevych, Yevgen  and
      Schneider, Nathan  and
      Bisazza, Arianna",
    editor = "Bonial, Claire  and
      Berzak, Yevgeni",
    booktitle = "Proceedings of the 30th Conference on Computational Natural Language Learning",
    month = jul,
    year = "2026",
    address = "San Diego, California, USA",
    publisher = "Association for Computational Linguistics",
    url = "https://aclanthology.org/2026.conll-main.23/",
    doi = "10.18653/v1/2026.conll-main.23",
    pages = "395--420",
    ISBN = "979-8-89176-410-1"
}

@inproceedings{reimers2019sentence,
    title = "Sentence-{BERT}: Sentence Embeddings using {S}iamese {BERT}-Networks",
    author = "Reimers, Nils and Gurevych, Iryna",
    booktitle = "Proceedings of the 2019 Conference on Empirical Methods in Natural Language Processing and the 9th International Joint Conference on Natural Language Processing (EMNLP-IJCNLP)",
    year = "2019",
    address = "Hong Kong, China",
    publisher = "Association for Computational Linguistics",
    url = "https://aclanthology.org/D19-1410",
}

@article{bunzeck2025richness,
  title={The richness of the stimulus: Constructional variation and development in child-directed speech},
  author={Bunzeck, Bastian and Diessel, Holger},
  journal={First Language},
  volume={45},
  number={2},
  pages={152--176},
  year={2025},
  publisher={Sage Publications Sage UK: London, England}
}

@article{rasanen2025pipeline,
  title={A pipeline for stochastic and controlled generation of realistic language input for simulating infant language acquisition},
  author={R{\"a}s{\"a}nen, Okko and Kocharov, Daniil},
  journal={Behavior Research Methods},
  volume={57},
  number={10},
  pages={275},
  year={2025},
  publisher={Springer}
}

@inproceedings{valentini2023automatic,
  title={On the automatic generation and simplification of children’s stories},
  author={Valentini, Maria and Weber, Jennifer and Salcido, Jesus and Wright, T{\'e}a and Colunga, Eliana and von der Wense, Katharina},
  booktitle={Proceedings of the 2023 Conference on Empirical Methods in Natural Language Processing},
  pages={3588--3598},
  year={2023}
}

@article{brysbaert2014concreteness,
  title={Concreteness ratings for 40 thousand generally known English word lemmas},
  author={Brysbaert, Marc and Warriner, Amy Beth and Kuperman, Victor},
  journal={Behavior research methods},
  volume={46},
  number={3},
  pages={904--911},
  year={2014},
  publisher={Springer}
}

@inproceedings{guo2024curious,
  title={The curious decline of linguistic diversity: Training language models on synthetic text},
  author={Guo, Yanzhu and Shang, Guokan and Vazirgiannis, Michalis and Clavel, Chlo{\'e}},
  booktitle={Findings of the Association for Computational Linguistics: NAACL 2024},
  pages={3589--3604},
  year={2024}
}

@article{dawson2021features,
  title={Features of lexical richness in children’s books: Comparisons with child-directed speech},
  author={Dawson, Nicola and Hsiao, Yaling and Banerji, Nilanjana and Tan, Alvin Wei M and Nation, Kate},
  journal={Language development research},
  year={2021},
  publisher={Carnegie Mellon University Library Publishing Service}
}

@article{duran2019align,
  title={ALIGN: Analyzing linguistic interactions with generalizable techNiques—A Python library.},
  author={Duran, Nicholas D and Paxton, Alexandra and Fusaroli, Riccardo},
  journal={Psychological methods},
  volume={24},
  number={4},
  pages={419},
  year={2019},
  publisher={American Psychological Association}
}

@article{cameron2003construction,
  title={A construction based analysis of child directed speech},
  author={Cameron-Faulkner, Thea and Lieven, Elena and Tomasello, Michael},
  journal={Cognitive science},
  volume={27},
  number={6},
  pages={843--873},
  year={2003},
  publisher={Wiley Online Library}
}

\appendix
\onecolumn
\section{Prompts for generation}
\label{sec:prompts}

The following dialogue summaries were used as seeds for synthetic data generation from our own model.

\begin{itemize}
    \item The child wants to watch paw patrol and play with her paw patrol toys, while the mother is cleaning the kitchen. The child then asks for cornflakes that look like paw patrol characters.
    \item Mother and child are in the car. The kid giggles because it got its icecream everywhere, mother tries to be strict but then also has to laugh.
    \item Kid wants to play Minecraft on the ipad but is not allowed until it puts its toys away. Mother turns it into a game with a song.
    \item Kid refuses to put on shoes because they are too scratchy. Mother negotiates by letting the child pick between two pairs.
    \item The child found a snail in the garden and wants to keep it as a pet. Mother explains snails live outside, child names it Lucas.
    \item Bath time. The kid pours water over the edge of the tub playing "waterfall". Mother tries to stay calm while mopping up.
    \item The child asks where the moon goes during the day. Mother gives a silly answer first, then tries to explain properly that it is behind the earth.
    \item Mother and child are folding laundry. The kid puts socks on its hands and pretends they are puppets talking to each other.
    \item Child is missing its favourite stuffed toy, which they left at her grandmas house. Mother calms her by reading a story.
    \item Child does not want to eat broccoli and hides it under the plate. Mother pretends the broccoli are little trees for a hungry dinosaur.
    \item Mother and child are baking cookies. The kid keeps sneaking bites of dough and blames it on the teddy bear sitting on the counter.
    \item The child does not want to go to bed and claims it is not tired, then yawns mid-sentence. Mother offers one more story as a compromise.
    \item At the supermarket the kid wants to push the big cart alone and bumps into a shelf. Mother suggests they steer it together like a race car.
    \item The child drew on the wall with crayons and proudly shows mother the "beautiful picture". Mother explains paper is for drawing and they clean it up together.
    \item It is raining and the child wants to jump in puddles. Mother says only with rain boots on, and they count the puddles on the way to the park.
    \item The kid finds mother\'s old phone in a drawer and asks why it has buttons. Mother explains phones looked different when she was little.
    \item The child is scared of a shadow on the bedroom wall. Mother shows that it is just the lamp and the curtain, and they make shadow animals together.
    \item Mother is watering the plants and the child wants to help, but pours too much water into one pot. Mother explains plants only drink a little bit.
    \item The child insists on wearing a superhero cape to kindergarten. Mother agrees it can come along but has to wait in the backpack during circle time.
    \item The kid is learning to zip up its own jacket and gets frustrated when the zipper sticks. Mother shows a trick and cheers when it finally works.
    \item The child hears the ice cream truck outside and begs to go, but dinner is almost ready. Mother promises ice cream for dessert on the weekend.
    \item Mother and child plant sunflower seeds in a pot. The kid wants to dig them up an hour later to check if they have grown yet.
    \item The child has the hiccups and finds it hilarious. Mother suggests silly cures like drinking water upside down and holding their breath.
    \item The kid builds a huge tower of blocks and it falls over just before finishing. Mother comforts the child and they build an even bigger one together.
    \item On a walk the child wants to bring home every stick and stone it finds. Mother says it can choose three favourites to put on the windowsill.
\end{itemize}
\end{document}